\documentclass[lettersize,journal]{IEEEtran}
\usepackage{amsmath,amsfonts}
\usepackage{algorithmic}
\usepackage{algorithm}
\usepackage{array}
\usepackage[caption=false,font=normalsize,labelfont=sf,textfont=sf]{subfig}
\usepackage{textcomp}
\usepackage{stfloats}
\usepackage{url}
\usepackage{verbatim}
\usepackage{graphicx}
\usepackage{amssymb}
\usepackage{cite}
\usepackage{placeins}
\begin{document}

\title{Empirical Investigation of the Impact of Phase Information on Fault Diagnosis of Rotating Machinery}

\author {
    Hiroyoshi Nagahama, 
    Katsufumi Inoue, 
    Masayoshi Todorokihara, and
    Michifumi Yoshioka
    \thanks{H. Nagahama, K. Inoue, and M. Yoshioka are with Graduate School of Informatics, Osaka Metropolitan University, Osaka Japan. }
    \thanks{M. Todorokihara is with Microdevices Operations Division, Seiko Epson Corp., Nagano, Japan. }
    
}



\maketitle

\begin{abstract}
Predictive maintenance of rotating machinery increasingly relies on vibration signals, yet most learning-based approaches either discard phase during spectral feature extraction or use raw time-waveforms without explicitly leveraging phase information. This paper introduces two phase-aware preprocessing strategies to address random phase variations in multi-axis vibration data: 
(1) three-axis independent phase adjustment that aligns each axis individually to zero phase
(2) single-axis reference phase adjustment that preserves inter-axis relationships by applying uniform time shifts.
Using a newly constructed rotor dataset acquired with a synchronized three-axis sensor, we evaluate six deep learning architectures under a two-stage learning framework. Results demonstrate architecture-independent improvements: the three-axis independent method achieves consistent gains (+2.7\% for Transformer), while the single-axis reference approach delivers superior performance with up to 96.2\% accuracy (+5.4\%) by preserving spatial phase relationships. These findings establish both phase alignment strategies as practical and scalable enhancements for predictive maintenance systems.
\end{abstract}

\begin{IEEEkeywords}
Anomaly Detection, Phase Adjustment, Deep Learning, Three-axis Synchronized Sensor 
\end{IEEEkeywords}

\section{Introduction}
\IEEEPARstart{M}{aintenance} plays a pivotal role in industrial operations because any unplanned downtime can severely disrupt production schedules and degrade overall business performance.
Maintenance strategies are generally classified into three categories: reactive (corrective), preventive, and predictive approaches~\cite{Mobley2002, Swanson2001, Gertsbakh2013, WanTII2017, Nguyen2015, WangJIM2017, CachadaMaintenance40, LiIndustry40, OkaforEFNMS2016, WangWIT2016}.

Reactive maintenance restores functionality after a failure occurs, while preventive maintenance schedules interventions based on mean time between failures, which often results in unnecessary maintenance and increased costs.
Predictive maintenance, in contrast, enables online condition monitoring and timely intervention before failure, striking a balance between cost and reliability. With the evolution of modern sensing technologies, predictive maintenance has become increasingly important for reducing maintenance frequency and costs.

Among various sensing modalities—such as temperature, pressure, electrical current, voltage, sound, and vibration—vibration signals are particularly effective for rotating machinery because they capture subtle changes and early signs of defects across diverse failure modes. Consequently, vibration-based monitoring has been widely adopted in fault diagnosis.

In practice, vibration data are typically acquired using three-axis acceleration sensors (e.g., IMUs), and displacement is often derived through double integration of acceleration signals before feature extraction. Since integration inherently amplifies low-frequency errors and sensor drift, even minor inaccuracies in the raw acceleration data can produce substantial displacement errors. This process highlights the importance of accurate sensing and preprocessing.

Recent advances in machine learning, especially deep learning, have significantly improved time-series analysis. Convolutional Neural Network (CNN)-based methods~\cite{NagahamaDCAI2024, HojoDCAI2025, TimesNet} and Transformer-based architectures~\cite{NegiAROB2024, Autoformer, FEDformer, iTransformer, TimeXer, NonStationary} have demonstrated strong performance when applied to vibration data~\cite{LiuMSSP2018, SurveyMaintenance}. Despite these advances, practical deployment remains challenging, leaving room for improvement.

To enhance diagnostic performance, two complementary directions can be considered: (i) developing more sophisticated models and (ii) improving the discriminative power of input representations. As numerous model architectures have already been proposed, this study focuses on the latter—enhancing feature representation through refined sensing and preprocessing.

One promising direction is to exploit phase information in vibration signals. According to ISO 20816-1~\cite{ISO20816}, phase information can reveal subtle changes that amplitude-only features fail to capture, supporting early detection of defects. However, phase has been largely ignored in existing approaches due to limitations in sensor performance and difficulties in achieving precise synchronization across axes. Conventional acceleration sensors often provide unreliable phase measurements, and inter-axis synchronization is problematic. Fortunately, high-precision three-axis vibration sensors with excellent synchronization characteristics have recently become available, enabling accurate phase acquisition.

Building on this capability, our research revisits vibration-based fault diagnosis and focuses on two critical factors: (i) synchronized multi-axis sensing and (ii) phase-consistent determination of segmentation onset for feature extraction. In common practice, long sequences are segmented into short windows for computational efficiency. When segmentation onset is chosen arbitrarily, identical machine states yield inconsistent phase representations, scattering features and blurring class boundaries. While amplitude-only features are relatively insensitive to onset, phase-sensitive features are strongly affected by onset ambiguity. To address this, we propose two contrasting preprocessing methods for phase alignment based on the dominant frequency phase.
The first, a three-axis independent method, suppresses random phase variation on each axis individually.
The second, a single-axis reference method, not only removes random phase variation but also preserves inter-axis spatial vibration patterns, which are essential for multi-axis recognition.

To the best of our knowledge, this is the first empirical study to explicitly leverage synchronized multi-axis phase information for machinery fault diagnosis using a high-precision three-axis sensor and to systematically compare phase-handling strategies across modern architectures. Our contributions are threefold:
\begin{itemize}
\item We demonstrate the impact of phase-aware preprocessing by exploiting synchronized three-axis vibration sensing and show that preserving inter-axis relative phase improves discriminability.
\item We propose a simple, reproducible criterion for phase-consistent segmentation onset and validate its effectiveness across CNN- and Transformer-based models.
\item We construct and release a multi-condition rotating machinery dataset acquired with the aforementioned sensor to support reproducible research (the dataset is available on \textit{\url{https://www.epsondevice.com/sensing/en/dataset}}).
\end{itemize}

\section{Related Studies}
Fault diagnosis of machinery is a traditional problem to prevent any unplanned downtime of machinery. 
To address this issue, numerous researchers have proposed machine learning-based methods including Decision Tree~\cite{Abdallah2018, BenkerchaSE2018, KouJRRT2018, LiATE2018, Patil2019}, 
Support Vector Machine (SVM)~\cite{HanATE2019, SantosSonsors2015, SoualhiTIM2015, SunATE2016, ZhuAccess2020}, 
k-Nearest Neighbor ($k$-NN)~\cite{AppanaACALCI2017, BaraldiEAAI2016, MadetiSE2018, TianTIE2016, SharifSV2016, XiongSJ2016}, 
Particle Filter~\cite{DarooghehCST2018}, Principal Component Analysis (PCA)~\cite{DengNNLS2018}, 
Self-Organizing Map~\cite{RaiJMES2018}. 
Additionally, since the advancement of deep learning-based approaches, recently, 
Auto Encoder (AE)-based methods~\cite{JiaNC2018, ShaoMSSP2017, HaidongKBS2018, LvJC2017, MaoTIM2019},
Deep Brief Network (DBN)-based methods~\cite{ChenTIM2017, ShaoISAT2017, WangMSSP2018, ZhuMLICOM2019, TimeMixer}, 
CNN-based methods~\cite{NagahamaDCAI2024, HojoDCAI2025, TimesNet, JiaACCESS2019, LiSensor2019, LiuSDPC2017, ZhangMSSP2025, AlqununSR2025}, 
Long-Short Term Memory (LSTM)-based methods~\cite{OnishiAROB2023, NoussisIFAC2024}, and 
the combination methods among them~\cite{FuMBE2024, KalayMachiens2025, WuCMC2024, TavaresIECON2024} 
can provide an accurate time-series analysis. 
Moreover, after the proposal of Transformer~\cite{Transformer}, 
many time-series analysis methods based on Transformers have been proposed~\cite{NegiAROB2024, Autoformer, FEDformer, iTransformer, TimeXer, NonStationary} and these outperform the competitive methods. 

As we can see from the above related studies, various approaches are being proposed daily.
However, towards the realization of precise predictive maintenance, 
these have not yet reached a practical level. 
To address this limitation, we revisit the signal information (vibration one in this research) for fault diagnosis. 
As mentioned in the previous section, the above related studies employ the amplitude information extracted from signal data with Fast Fourier Transform (FFT), Wavelet transform, etc., 
or directly leverage the signal data captured from sensors as is. 
In the former case, the phase information of signal data, which can support the detection of subtle changes based on defects of machinery, is discarded. 
Besides, in the latter case, the phase information is not explicitly utilized. 
Therefore, we exploit the phase information in this research by using a three-axis vibration sensor with excellent synchronization characteristics. 
This is the main difference from the related studies. Based on the phase information, in this research, we propose a new pre-processing method to enhance the signal data for accurate fault classification. 

\section{Phase Adjustment Approaches for Multi-axis Vibration Data}

In this section, we propose phase adjustment approaches for multi-axis 
vibration data to address the segmentation onset determination problem. 
We first present a phase-based criterion for determining the segmentation 
onset (Section III-A), then formulate two phase adjustment 
methods (Section III-B), and finally compare their characteristics (Section III-C).

The segmentation onset problem arises from the nature of continuous machine monitoring. For fault diagnosis, machines are continuously or periodically monitored with sensors. For example, vibration sensors mounted on rotating machinery capture vibration information at regular intervals. Such captured information forms long-term time-series data utilized for fault diagnosis. Due to computational complexity, long-term data must be sequentially segmented into shorter windows—an inevitable pre-processing step before feature extraction. However, this segmentation process harbors a critical issue: the starting point for segmentation is typically chosen arbitrarily from the continuous data stream.

This arbitrary choice of segmentation onset creates a fundamental problem: 
signal features extracted from identical machine conditions exhibit random phase variations, causing feature vectors to be scattered across the feature space and creating ambiguous boundaries between different conditions.
The random segmentation starting point degrades feature consistency and adversely affects fault diagnosis performance. Therefore, determining proper segmentation onset is crucial for achieving accurate fault diagnosis.


Recent advances in sensor technology have enabled the use of three-axis vibration sensors with excellent synchronization characteristics, which can precisely capture detailed phase information across multiple axes. This enhanced capability motivates us to develop new signal processing methods that fully exploit this rich phase information.

Specifically, we propose a phase-based criterion to determine the optimal segmentation onset and corresponding pre-processing methods to align signal segments based on phase relationships. These approaches enable the extraction of consistent signal features regardless of the arbitrary starting point, thereby ensuring feature consistency across identical machine conditions.
In the following sections, we explain the proposed method in more detail.


\subsection{Determination Criterion of Segmentation Onset}
In this section, we explain the concrete process of determining the onset of the segmentations of the signal data.


\begin{figure}[t]
    \centering
    \includegraphics[width=\columnwidth]{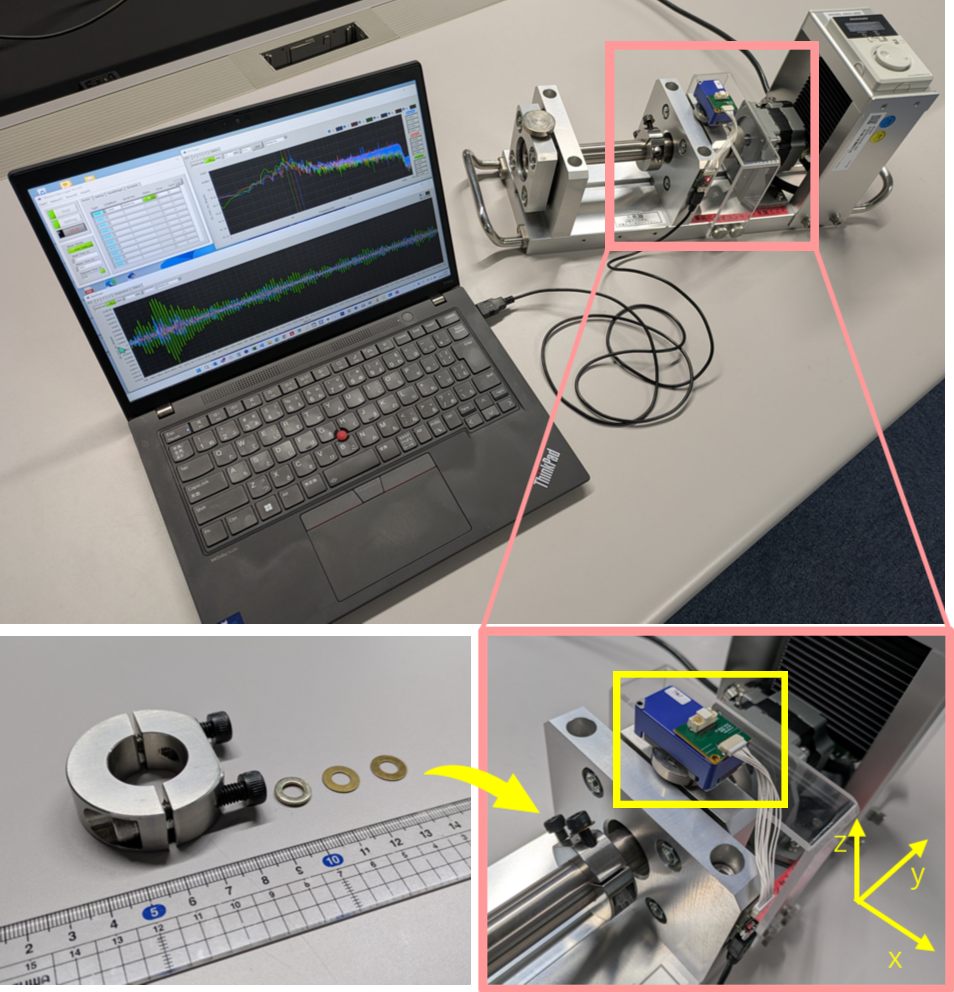}
    \caption{Experimental setup for vibration data acquisition. The rotorkit apparatus features a three-axis vibration sensor mounted on the bearing housing with coordinate system (X: lateral, Y: axial, Z: vertical). Controlled fault conditions are generated using washer-based unbalance configurations. The bottom left shows the washers used: thick washer (left) and thin washers (center and right) for creating distinct anomaly classes.}
    \label{fig:sensor_setup}
\end{figure}

In general, a three-axis vibration sensor is mounted based on the characteristics of a monitored machine. For rotating machinery, as illustrated in Fig.\ref{fig:sensor_setup}, the coordinate system is defined with the Z-axis along the gravitational direction, the Y-axis along the rotational axis, and the X-axis perpendicular to both (forming a right-handed coordinate system). In rotating machinery, common fault modes such as unbalance and bearing defects manifest primarily as radial vibrations (perpendicular to the rotation axis), while axial vibrations are typically smaller. Based on this characteristic, we designate the X-axis (radial direction) as the reference axis for our phase-based segmentation criterion. As theoretically shown in Section III-C and empirically validated in Section IV-B, the method maintains equivalent performance regardless of reference axis selection due to preserved relative phase relationships.

In addition, a normal rotating machine is rotating at a constant speed. 
Therefore, this rotating frequency indicates the strong intensity in the frequency analysis. 
From this fact, this frequency is regarded as a ``\textit{dominant frequency}" and 
is leveraged to determine the segmentation onset. The dominant frequency can be either directly specified from known operational parameters (e.g., rotational speed in RPM) or automatically detected via FFT analysis of an initial segment. 
To determine the segmentation onset, we evaluate the phase information at this dominant frequency in the vibration data on the reference axis and determine a timing such that the phase information becomes   $\phi$[rad] ($\phi = 0$ in this research) as the segmentation onset.
From this process, the onset is consistently determined independently of the onset of machine monitoring. 

\subsection{Mathematical Formulation of Phase Adjustment}

This section presents the mathematical formulation of phase adjustment methods for multi-axis vibration data. We define the signal processing operations for both the three-axis independent adjustment and the proposed single-axis reference adjustment approach.

Consider a three-axis vibration sensor that captures synchronized signals from a rotating machine. Let the raw vibration signals be denoted as:
\begin{equation}
\mathbf{\textit{s}}(t) = [x(t), y(t), z(t)]^\top
\end{equation}
where $x(t)$, $y(t)$, and $z(t)$ represent the time-domain signals along the X, Y, and Z axes, respectively.

For analysis, we extract finite-length segments from the continuous data stream. The $i$-th segment is denoted as:
\begin{equation}
\mathbf{\textit{s}}^{(i)}(t) = [x^{(i)}(t), y^{(i)}(t), z^{(i)}(t)]^\top, \quad t \in [0, L-1]
\end{equation}
where the superscript $(i)$ denotes the $i$-th segment and $L$ represents the number of samples in the segment.

The discrete Fourier transform (DFT) of each axis signal is computed as:
\begin{align}
X^{(i)}[k] &= \sum_{n=0}^{L-1} x^{(i)}(nT_s) e^{-j2\pi kn/L} \\
Y^{(i)}[k] &= \sum_{n=0}^{L-1} y^{(i)}(nT_s) e^{-j2\pi kn/L} \\
Z^{(i)}[k] &= \sum_{n=0}^{L-1} z^{(i)}(nT_s) e^{-j2\pi kn/L}
\end{align}
where $T_s$ is the sampling period, and $k$ is the frequency bin index. The phase information at the dominant frequency $f_d$ (corresponding to bin index $k_d$) is extracted as:
\begin{align}
\phi^{(i)}_x &= \arg(X^{(i)}[k_d]) \\ 
\phi^{(i)}_y &= \arg(Y^{(i)}[k_d]) \\
\phi^{(i)}_z &= \arg(Z^{(i)}[k_d]) 
\end{align}
where $\arg(\cdot)$ denotes the argument (phase angle) of a complex number.

\subsubsection{Three-axis Independent Phase Adjustment}
\indent As a first approach to achieve phase consistency across signal segments, 
we propose the three-axis independent phase adjustment method. In this 
method, each axis is processed separately to align its phase at the 
dominant frequency to zero.
The required time shift to achieve zero phase is calculated for each axis, respectively:

\begin{equation}
\Delta t^{(i)}_x = -\frac{\phi^{(i)}_x}{2\pi f_d}, \quad \Delta t^{(i)}_y = -\frac{\phi^{(i)}_y}{2\pi f_d}, \quad \Delta t^{(i)}_z = -\frac{\phi^{(i)}_z}{2\pi f_d}
\end{equation}
Then, the adjusted signals $x'^{(i)}(t),y'^{(i)}(t),z'^{(i)}(t)$ are obtained by applying the respective time shifts:
\begin{align}
x'^{(i)}(t) &= x^{(i)}(t + \Delta t^{(i)}_x) \\
y'^{(i)}(t) &= y^{(i)}(t + \Delta t^{(i)}_y) \\
z'^{(i)}(t) &= z^{(i)}(t + \Delta t^{(i)}_z)
\end{align}
After adjustment, as shown in the middle panel of Fig.~\ref{fig:phase_adjustment_comparison}, the phase at the dominant frequency $\phi'^{(i)}_x, \phi'^{(i)}_y, \phi'^{(i)}_z$ becomes zero for all axes:
\begin{equation}
\phi'^{(i)}_x = \phi'^{(i)}_y = \phi'^{(i)}_z = 0
\end{equation}
While this method ensures phase consistency for each axis individually, the relative phase relationships between axes are lost:
\begin{equation}
\phi'^{(i)}_{xy} = \phi'^{(i)}_x - \phi'^{(i)}_y = 0 - 0 = 0
\end{equation}

\begin{equation}
\therefore \quad \phi'^{(i)}_{xy} \neq \phi^{(i)}_{xy}
\end{equation}

This loss of inter-axis phase information eliminates valuable spatial characteristics of the vibration patterns.
While this method achieves phase consistency for each axis, it destroys the inter-axis phase relationships, which contain important spatial information about fault patterns. This limitation motivates our second proposed approach that preserves these relationships.

\subsubsection{Single-axis Reference Phase Adjustment}
\indent In the single-axis reference approach, we use one axis as the reference to determine the time shift applied uniformly to all three axes. This design choice preserves the inter-axis phase relationships while achieving phase consistency.

The choice of reference axis depends on the sensor mounting configuration and the specific characteristics of the monitored machinery. Without loss of generality, we denote the reference axis as the X-axis in the following formulation. The required time shift is calculated based solely on the reference axis phase:

\begin{equation}
\Delta t^{(i)} = -\frac{\phi^{(i)}_x}{2\pi f_d}
\end{equation}
where $\phi^{(i)}_x$ is the phase of the X-axis at the dominant frequency 
$f_d$ for the $i$-th segment. This time shift brings the reference axis 
phase to zero.

Then, we apply the identical time shift to all three axes:
\begin{align}
x''^{(i)}(t) &= x^{(i)}(t + \Delta t^{(i)}) \\
y''^{(i)}(t) &= y^{(i)}(t + \Delta t^{(i)}) \\
z''^{(i)}(t) &= z^{(i)}(t + \Delta t^{(i)})
\end{align}

After applying the uniform time shift, the phases at the dominant frequency become:
\begin{align}
\phi''^{(i)}_x &= \phi^{(i)}_x - \Delta\phi \\
\phi''^{(i)}_y &= \phi^{(i)}_y - \Delta\phi \\
\phi''^{(i)}_z &= \phi^{(i)}_z - \Delta\phi
\end{align}
where $\Delta\phi = 2\pi f_d \Delta t^{(i)}$ is the phase shift applied equally to all axes.

As shown in the right panel of Fig.~\ref{fig:phase_adjustment_comparison}, by applying the same time shift to all axes, the method ensures that the relative phase relationships are completely preserved:
\begin{equation}
\label{equ: preservation of inter-axis phase information}
\begin{aligned}
\phi''^{(i)}_{xy} &= \phi''^{(i)}_x - \phi''^{(i)}_y \\
&= (\phi^{(i)}_x - \Delta\phi) - (\phi^{(i)}_y - \Delta\phi) \\
&= \phi^{(i)}_x - \phi^{(i)}_y = \phi^{(i)}_{xy}
\end{aligned}
\end{equation}

This invariance property holds for all axis pairs, meaning that the spatial structure of the multi-axis vibration pattern is maintained.

This mathematical formulation demonstrates two key properties:

\begin{enumerate}
    \item reference axis normalization with $\phi''^{(i)}_x = 0$
    \item relative phase preservation with $\phi''^{(i)}_{jk} = \phi^{(i)}_{jk}$ for all axis pairs $(j,k)$
\end{enumerate}

The proposed method thus achieves both phase consistency (elimination of random variations) and spatial information preservation (maintenance of inter-axis relationships), addressing the fundamental limitations of the three-axis independent approach.

\subsection{Comparative Analysis of Phase Adjustment Methods}

To evaluate different phase adjustment approaches, we compare how each method preserves the critical information contained in the vibration data. Table~\ref{tab:phase_adjustment_comparison} summarizes this comparison across three key information types.

\begin{table}[t]
\centering
\caption{Comparison of Phase Adjustment Methods from Vibration Information Preservation Perspective}
\label{tab:phase_adjustment_comparison}
\begin{tabular}{l|c|c|c}
\hline
\textbf{Method} & \textbf{Amplitude} & \textbf{Phase} & \textbf{Relative phase} \\
\hline
No Adjustment & \checkmark Preserved & Random & \checkmark Preserved \\
\hline
Three-axis Independent & \checkmark Preserved & \checkmark Unified & Lost \\
\hline
\textbf{Single-axis Reference} & \textbf{\checkmark Preserved} & \textbf{\checkmark Unified} & \textbf{\checkmark Preserved} \\
\hline
\end{tabular}
\end{table}

\begin{figure*}[ht]
    \centering
    \includegraphics[width=\textwidth]{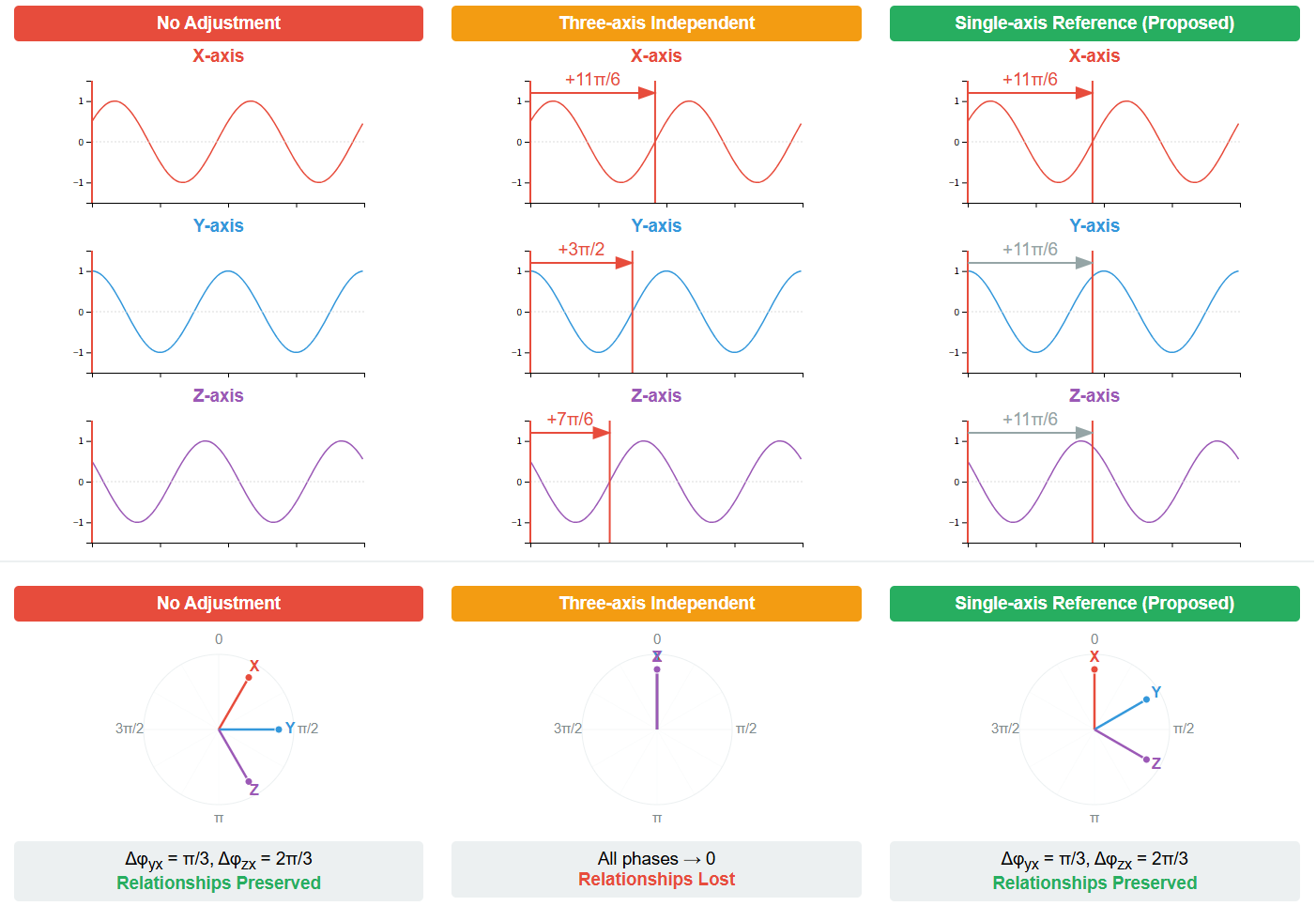}
    \caption{Comparison of phase adjustment methods for multi-axis vibration data. The upper panel shows time-domain signals for each axis with arrows indicating the time shift applied to each axis. Vertical lines indicate segmentation positions. The lower panel shows corresponding phase relationships in polar coordinates, demonstrating how each method affects inter-axis phase relationships.}
    \label{fig:phase_adjustment_comparison}
\end{figure*}

The three information types represent distinct aspects of vibration signals: amplitude information contains fault-related magnitude characteristics, phase information ensures learning consistency by eliminating random variations, and inter-axis relative phase preserves spatial vibration patterns essential for 3D fault recognition. As Table~\ref{tab:phase_adjustment_comparison} demonstrates, only the proposed single-axis reference method preserves all three types simultaneously.

To illustrate the fundamental differences between phase adjustment methods, Fig.~\ref{fig:phase_adjustment_comparison} presents a comprehensive comparison of the three approaches using both time-domain signals and phase vector representations.

Since the core of the proposed method lies in preserving inter-axis relative phase relationships, theoretically equivalent classification performance is expected regardless of which axis is selected as the reference. The relative phase differences between axes, $\phi^{(i)}_{jk} = \phi^{(i)}_j - \phi^{(i)}_k$, remain invariant under reference axis changes as demonstrated in Equation~\ref{equ: preservation of inter-axis phase information}. This invariance ensures that the spatial vibration patterns critical for anomaly detection are preserved regardless of reference selection, enabling comparable learning outcomes across different reference choices.

This theoretical prediction will be empirically validated in Section IV-B, where we demonstrate consistent performance improvements across X, Y, and Z axis selections as the reference.

\section{EXPERIMENTAL VALIDATION}

To comprehensively validate the effectiveness of our proposed single-axis reference phase adjustment method, we conducted systematic experiments using real vibration data from rotating machinery. Our experimental validation was designed with a two-fold objective: first, to demonstrate the progressive performance improvement achieved by our phase adjustment approach, and second, to verify the generalizability of this improvement across diverse deep learning architectures.

\subsection{Dataset and Experimental setup}


For vibration data acquisition, we employed a controlled rotorkit apparatus (Shinkawa Electric AA31-020) operating at 1200 RPM (20 Hz fundamental frequency) with 3 kHz sampling rate. A three-axis digital quartz vibration sensor (Seiko Epson M-A342) was mounted on the upper bearing housing, with X-axis in the lateral direction, Y-axis in the axial direction, and Z-axis in the vertical direction. This sensor achieves flat frequency response characteristics in the 10-1000 Hz operating bandwidth and excellent synchronization accuracy (less than 10 µs) through digital signal processing of three single-axis sensors with identical characteristics.

\begin{figure}[t]
    \centering
    \includegraphics[width=\columnwidth]{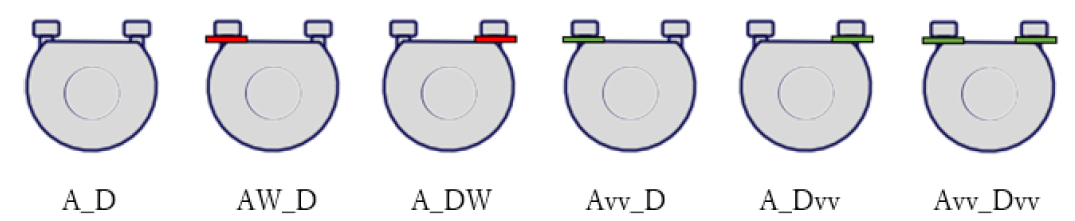}
    \caption{Washer Attachment Status and Label Names for Each Class (Red indicates thick washers, green indicates thin washers)}
    \label{fig:sensor_detailed_config}
\end{figure}

Anomaly conditions were systematically created using washer-based unbalance configurations mounted on the rotating shaft. As illustrated in Fig.~\ref{fig:sensor_detailed_config}, we employed two types of washers with different masses: thick washers and thin washers, attached at two mounting positions along the shaft (left position and right position as viewed in the apparatus setup).

As shown in Fig.\ref{fig:sensor_detailed_config}, the dataset consists of 6 classes defined by systematic washer configurations at two mounting positions. Table~\ref{tab:dataset_classes} summarizes the complete class definitions.

\begin{table}[htbp]
\centering
\caption{Dataset classification based on washer configurations}
\label{tab:dataset_classes}
\begin{tabular}{c|c|c|c}
\hline
\textbf{Class} & \textbf{Condition} & \textbf{Left Position} & \textbf{Right Position} \\
\hline
Class 0 & Normal & - & - \\
\hline
Class 1 & Anomaly & Thick & - \\
\hline
Class 2 & Anomaly & - & Thick \\
\hline
Class 3 & Anomaly & Thin & - \\
\hline
Class 4 & Anomaly & - & Thin \\
\hline
Class 5 & Anomaly & Thin & Thin \\
\hline
\end{tabular}
\end{table}

For each class, we measured 16 independent data files with the rotor operating at 1200 RPM, constructing a total dataset of 96 files. Each data file contains 300,000 data points (100 seconds) of raw vibration data sampled at 3 kHz.


For data preprocessing, we removed the first 30,000 points (10 seconds) since sensor responses at the beginning of measurement contain transient phenomena. Subsequently, we extracted 10,000 points (3.3 seconds) from the remaining data as analysis data. This extraction reduced deep learning model training time to a practical range. The extracted data underwent window segmentation with input length 512 points and skip length 32 points, generating approximately 300 windows per file.
The processed dataset was divided into training, validation, and testing sets with a ratio of 11:1:4, corresponding to $69\%$ training (66 files), $6\%$ validation (6 files), and $25\%$ testing (24 files).

\begin{figure*}[t]
    \centering
    \includegraphics[width=\textwidth]{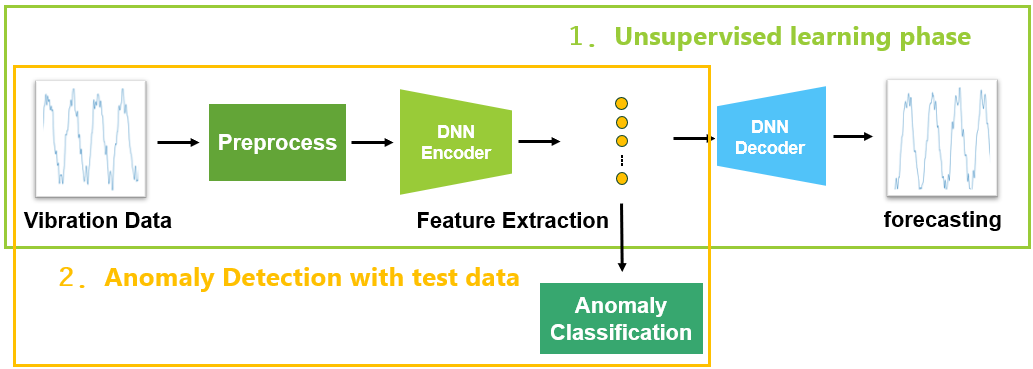}
    \caption{Two-stage experimental workflow for vibration anomaly detection.
    Stage 1 (Unsupervised learning phase): Deep learning models are trained on time series prediction tasks to extract meaningful temporal features from preprocessed vibration data. 
    Stage 2 (Anomaly detection with test data): Extracted encoder features are used for SVM-based anomaly classification.}
    \label{fig:Experimental Workflow}
\end{figure*}

Figure~\ref{fig:Experimental Workflow} illustrates the experimental workflow designed to validate our proposed phase adjustment method.
The two-stage learning framework employed in this study is based on the approach originally proposed by Onishi et al.~\cite{OnishiAROB2023} for vibration-based fault diagnosis using LSTM and feature orthogonalization. The experimental workflow consists of two sequential stages:
\begin{enumerate}
    \item unsupervised pre-training stage
    \item supervised classification stage
\end{enumerate}
that reflected realistic industrial deployment scenarios.

In the unsupervised pre-training stage, models performed time series prediction tasks with sequence length 512 and prediction length 3, learning to extract meaningful temporal features through forecasting without requiring anomaly labels. This stage enabled models to understand normal operational patterns from unlabeled historical data. 

Subsequently, in the supervised classification stage, SVM utilized the learned encoder features for anomaly detection. SVM hyperparameters ($\gamma$ and $C$) were optimized through comprehensive $3 \times 3$ grid search for each experimental condition to ensure fair comparison across different preprocessing methods.

This data set and experimental workflow explained above were utilized in all experiments given in this paper.

\subsection{Performance Analysis of Phase Adjustment Approaches}

To demonstrate the step-wise improvements of phase adjustment methods, we conducted detailed analysis using the two-stage experiment framework with Transformer encoder and decoder as DNN. We compared three preprocessing approaches to validate our proposed methods:
\begin{enumerate}
    \item \textbf{No phase adjustment} as the baseline condition
    \item \textbf{Three-axis independent phase adjustment} where each axis is individually aligned to zero phase
    \item \textbf{single-axis reference adjustment} with different reference axis selections (X, Y, Z axes) to verify robustness across axis choices
\end{enumerate}

Table~\ref{tab:progressive_analysis} presents the results of progressive performance analysis comparing all phase adjustment strategies.

\begin{table}[t]
\centering
\caption{Performance Analysis of Phase Adjustment Approaches (Transformer Model)}
\label{tab:progressive_analysis}
\begin{tabular}{l|c|c}
\hline
\textbf{preprocessing method} & \textbf{Accuracy(\%)$\pm$Std Dev} & \textbf{Improvement} \\
\hline
No Adjustment & 90.8$\pm$3.2 & Baseline \\
\hline
three Independent & 93.5$\pm$0.5 & +2.7pp \\
\hline
\textbf{X-axis Reference} & \textbf{96.2}\textbf{$\pm$1.1} & \textbf{+5.4pp} \\
\hline
Y-axis Reference & 95.2$\pm$1.3 & +4.4pp \\
\hline
Z-axis Reference & 96.1$\pm$0.5 & +5.3pp \\
\hline
\end{tabular}
\end{table}

The key experimental findings are as follows:

\textbf{Step-wise Enhancement Pattern}: 
The three-axis independent adjustment showed notable improvement from 90.8\% to 93.5\% (2.7 percentage point improvement) compared to the baseline without phase adjustment. As established in Section III, this improvement results from eliminating phase inconsistencies across windows, enabling more stable and consistent feature learning.

The single-axis reference approach achieved further improvement to 95.2–96.2\% accuracy across different reference axis selections, representing an additional 1.7-2.7 percentage point gain over the three-axis independent method. This additional improvement occurs because the single-axis reference method preserves inter-axis phase relationships, enabling the classifier to distinguish fault-specific three-dimensional vibration characteristics, while the three-axis independent approach eliminates these crucial inter-axis phase correlations.

\textbf{Reference Axis Robustness}: 
The consistent high performance across X/Y/Z axis selection (95.2-96.2\% range) confirms the theoretical prediction that preserved relative phase relationships, rather than specific reference choice, drive performance improvements. This robustness occurs because inter-axis phase information is preserved regardless of which axis serves as the reference, as demonstrated mathematically in Equation~\ref{equ: preservation of inter-axis phase information}.

\subsection{Generalizability Assessment across Model Architecture}

To demonstrate the generalizability of the proposed phase adjustment method across different architectures, we evaluated six diverse deep learning architectures representing different technological approaches to time series analysis:

\begin{enumerate}
    \item \textbf{Transformer}~\cite{Transformer}: Standard multi-head attention mechanism for sequence modeling
    \item \textbf{Autoformer}~\cite{Autoformer}: Transformer-based decomposition architecture with seasonal-trend decomposition
    \item \textbf{FEDformer}~\cite{FEDformer}: Transformer-based frequency enhanced decomposed architecture
    \item \textbf{TimesNet}~\cite{TimesNet}: CNN-based multi-periodicity modeling architecture
    \item \textbf{TimeXer}~\cite{TimeXer}: Transformer-based patch embedding approach for time series
    \item \textbf{Non-Stationary Transformer}~\cite{NonStationary}: Transformer with adaptive normalization for non-stationary time series
\end{enumerate}

This diverse architectural selection ensured that observed improvements were not dependent on specific learning mechanisms but represent fundamental benefits of preserved phase relationships.

For each architecture, we applied the same experimental protocol established in Section IV-A, comparing three preprocessing conditions: no phase adjustment, three-axis independent adjustment, and single-axis reference adjustment (X-axis). All models followed identical training procedures with the two-stage framework, maintaining consistent evaluation conditions across architectures to ensure fair comparison.

Table~\ref{tab:cross_architecture} demonstrates that phase adjustment methods are effective regardless of architecture.
\begin{table}[t]
\centering
\caption{Cross-Architecture Performance Validation}
\label{tab:cross_architecture}
\begin{tabular}{l|c|c|c}
\hline
\textbf{Model} & \textbf{No Adj.} & \textbf{Three-axis Ind.} & \textbf{Single-axis Ref.} \\
\hline
Transformer~\cite{Transformer} & 90.8$\pm$3.2 & 93.5$\pm$0.5 & \textbf{96.2$\pm$1.1} \\
\hline
TimesNet~\cite{TimesNet} & 88.9$\pm$2.2 & 94.8$\pm$0.2 & \textbf{96.6$\pm$0.3} \\
\hline
TimeXer~\cite{TimeXer} & 92.7$\pm$0.7 & 93.8$\pm$0.7 & \textbf{98.0$\pm$0.2} \\
\hline
Non-Stationary~\cite{NonStationary} & 88.3$\pm$1.2 & 93.3$\pm$0.6 & \textbf{95.0$\pm$1.2} \\
\hline
FEDformer~\cite{FEDformer} & 73.1$\pm$9.4 & \textbf{87.7$\pm$4.0} & 87.3$\pm$1.4 \\
\hline
Autoformer~\cite{Autoformer} & 75.8$\pm$0.1 & \textbf{90.3$\pm$1.7} & 90.2$\pm$0.2 \\
\hline
\end{tabular}
\end{table}
The experimental results revealed consistent improvement patterns across diverse architectures. Performance improvements through phase adjustment methods were confirmed across all six architectures. Four models (Transformer, TimesNet, TimeXer, Non-Stationary) achieved highest performance with the single-axis reference method, while two models (FEDformer, Autoformer) showed highest performance with the three-axis independent method. The effectiveness of phase adjustment was confirmed across diverse learning paradigms including attention mechanisms, frequency processing, decomposition approaches, and patch-based methods. Notably, FEDformer and Autoformer with the three -axis independent approach showed substantial improvements of 14.6 and 14.5 points, respectively. This was likely attributed to their lower baseline performance, which provided greater room for improvement.

\section{Conclusion \& Future Work}
This work revisited the role of phase information in vibration-based fault diagnosis of rotating machinery. By exploiting synchronized multi-axis acquisition, we showed that explicitly organizing the phase at the segment onset matters: the proposed \emph{single-axis reference phase adjustment} removed random phase variation while preserving inter-axis relative phase, thus stabilizing feature extraction and improving classification accuracy. Progressive analysis confirmed a clear hierarchy — no adjustment $<$ independent axis alignment $<$ single-axis reference alignment — and cross-architecture experiments demonstrated that the benefit was generic across Transformer variants and CNN-based sequence models. We also released the dataset and code to facilitate reproducibility and broader benchmarking (the dataset is available on \textit{\url{https://www.epsondevice.com/sensing/en/dataset}}).

In the future work, we plan to (i) extend the dataset to diverse fault modes (bearing defects, misalignment, looseness) and multiple speeds, (ii) study automatic dominant-frequency detection and multi-frequency phase synchronization for machines with multi-band or variable-speed operation, (iii) compare different reference-axis selection strategies and investigate adaptive reference selection, (iv) evaluate end-to-end phase-aware neural encoders that inherently model relative phase without external alignment, (v) conduct ablation on window length, hop size, and label granularity, and (vi) assess robustness across sensors and mounting configurations, including cross-device transfer.



\bibliographystyle{IEEEtran}
\bibliography{myref}

\begin{thebibliography}{1}
\bibliographystyle{IEEEtran}

\bibitem{ref1}
{\it{Mathematics Into Type}}. American Mathematical Society. [Online]. Available: https://www.ams.org/arc/styleguide/mit-2.pdf

\bibitem{ref2}
T. W. Chaundy, P. R. Barrett and C. Batey, {\it{The Printing of Mathematics}}. London, U.K., Oxford Univ. Press, 1954.

\bibitem{ref3}
F. Mittelbach and M. Goossens, {\it{The \LaTeX Companion}}, 2nd ed. Boston, MA, USA: Pearson, 2004.

\bibitem{ref4}
G. Gr\"atzer, {\it{More Math Into LaTeX}}, New York, NY, USA: Springer, 2007.

\bibitem{ref5}M. Letourneau and J. W. Sharp, {\it{AMS-StyleGuide-online.pdf,}} American Mathematical Society, Providence, RI, USA, [Online]. Available: http://www.ams.org/arc/styleguide/index.html

\bibitem{ref6}
H. Sira-Ramirez, ``On the sliding mode control of nonlinear systems,'' \textit{Syst. Control Lett.}, vol. 19, pp. 303--312, 1992.

\bibitem{ref7}
A. Levant, ``Exact differentiation of signals with unbounded higher derivatives,''  in \textit{Proc. 45th IEEE Conf. Decis.
Control}, San Diego, CA, USA, 2006, pp. 5585--5590. DOI: 10.1109/CDC.2006.377165.

\bibitem{ref8}
M. Fliess, C. Join, and H. Sira-Ramirez, ``Non-linear estimation is easy,'' \textit{Int. J. Model., Ident. Control}, vol. 4, no. 1, pp. 12--27, 2008.

\bibitem{ref9}
R. Ortega, A. Astolfi, G. Bastin, and H. Rodriguez, ``Stabilization of food-chain systems using a port-controlled Hamiltonian description,'' in \textit{Proc. Amer. Control Conf.}, Chicago, IL, USA,
2000, pp. 2245--2249.

\end{thebibliography}

\end{document}